%% file: templateArxiv.tex
\documentclass{article}

\usepackage{PRIMEarxiv}

\usepackage[utf8]{inputenc} 
\usepackage[T1]{fontenc}    
\usepackage{hyperref}       
\usepackage{url}            
\usepackage{booktabs}       
\usepackage{amsfonts}       
\usepackage{amsmath}       
\usepackage{nicefrac}       
\usepackage{microtype}      
\usepackage{lipsum}
\usepackage{fancyhdr}       
\usepackage{graphicx}       
\graphicspath{{media/}}     

\title{TAPESTRY: From Geometry to Appearance via Consistent Turntable Videos
}

\author{
Yan Zeng$^{1,2\star}$ \quad Haoran Jiang$^{1,2\star}$ \quad Kaixin Yao$^{1,2\star}$  \quad Qixuan Zhang$^{1,2\dagger}$ \quad Longwen Zhang$^{1,2\dagger}$ \And Lan Xu$^{1\ddagger}$ \quad Jingyi Yu$^{1\ddagger}$ \\
\\
  $^1$ShanghaiTech University\quad 
  $^2$Deemos Technology \\
\\
{\tt\small \{zengyan2024,jianghr2024,yaokx2023zhangqx1,zhanglw2,xulan1,yujingyi\}@shanghaitech.edu.cn} \\
}

\begin{document}
\maketitle


\footnotetext{$\star$ Equal contributions. $\dagger$ Project leader. $\ddagger$ Corresponding author. } 
\input{sec/0_abstract}

\input{sec/1_intro}

\input{sec/2_related_work}

\input{sec/3_method}

\input{sec/4_experiments}
\input{sec/5_discussion}

\bibliographystyle{unsrt}  
\bibliography{references}

\end{document}

%% file: sec/0_abstract.tex
\begin{abstract}
Automatically generating photorealistic and self-consistent appearances for untextured 3D models is a critical challenge in digital content creation. The advancement of large-scale video generation models offers a natural approach: directly synthesizing 360-degree turntable videos, which can serve not only as high-quality dynamic previews but also as an intermediate representation to drive texture synthesis and neural rendering. However, existing general-purpose video diffusion models struggle to maintain strict geometric consistency and appearance stability across the full range of views, making their outputs ill-suited for high-quality 3D reconstruction.
To this end, we introduce TAPESTRY, a framework for generating high-fidelity turntable videos conditioned on explicit 3D geometry. We reframe the 3D appearance generation task as a geometry-conditioned video diffusion problem: given a 3D mesh, we first render and encode multi-modal geometric features to constrain the video generation process with pixel-level precision, thereby enabling the creation of high-quality and consistent turntable videos. Building upon this, we also design a method for downstream reconstruction tasks from the TTV input, featuring a multi-stage pipeline with 3D-Aware Inpainting. By rotating the model and performing a context-aware secondary generation, this pipeline effectively completes self-occluded regions to achieve full surface coverage.
The videos generated by TAPESTRY are not only high-quality dynamic previews but also serve as a reliable, 3D-aware intermediate representation that can be seamlessly back-projected into UV textures or used to supervise neural rendering methods like Gaussian Splatting. This enables the automated creation of production-ready, complete 3D assets from untextured meshes. Experimental results demonstrate that our method significantly outperforms existing approaches in both video consistency and final reconstruction quality.

\keywords{Video Generation \and 3D Texturing \and Geometric Consistency}

\end{abstract}

%% file: sec/1_intro.tex
\section{Introduction}

\begin{figure}[t]
\begin{center}
    \includegraphics[width=0.95\linewidth]{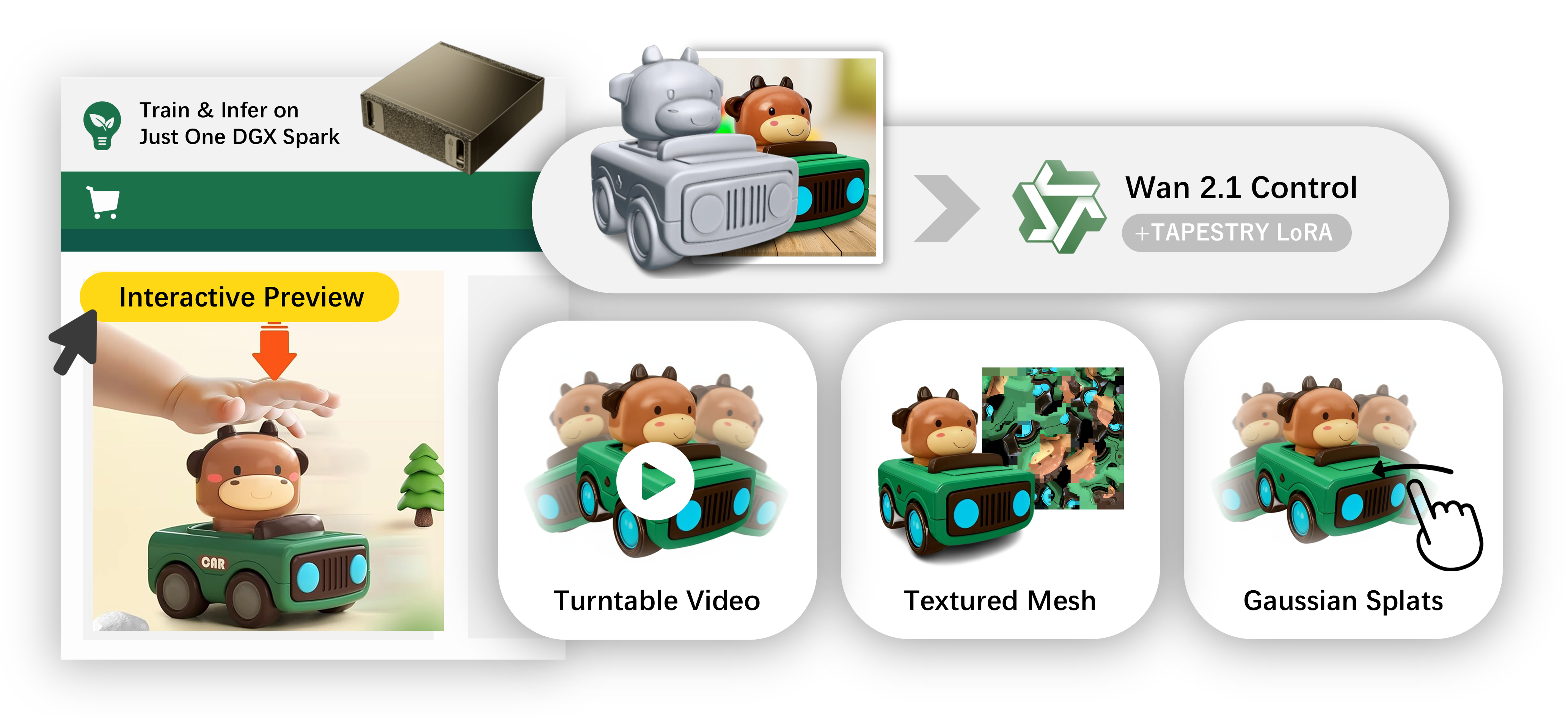}
  \caption{\small{We introduce TAPESTRY for high-fidelity 3D appearance generation by synthesizing a reconstructable Turntable Video through strong geometric conditioning. This highly consistent video then serves as a robust data source to create a final, high-quality asset. }}
  \label{fig:teaser}
  \end{center}
\end{figure}

Assigning realistic, physically plausible appearances to 3D models is a core step in digital content creation. Traditional workflows rely heavily on artists manually authoring textures, which is time-consuming and labor-intensive, requires both advanced artistic skills and a deep understanding of lighting and material physics, and does not scale to large asset libraries. Consequently, automatically generating high-quality, 3D-consistent appearances for untextured white 3D models remains a long-standing goal for both academia and industry.

With the rapid advancement of large-scale video generation models~\cite{yang2024cogvideox,kong2024hunyuanvideo,wan2025}, a new possibility has emerged: creating 360-degree turntable videos (TTVs) with generative AI using video diffusion models. In e-commerce, this format has become an industry standard, allowing consumers to inspect a product’s fit, fabric drape, texture, stitching, and finish with near in-store fidelity. It is used by luxury fashion houses such as Burberry to showcase garment flow and by major online marketplaces such as Amazon and Alibaba to reduce return rates and enhance buyer confidence. Beyond strictly serving as visual previews, high-fidelity TTVs function as a versatile dual-purpose asset. They provide an immediate, frictionless 3D-like experience on standard video platforms (e.g., social media feeds) where interactive rendering is unavailable, while simultaneously containing sufficient geometric and view signal to support advanced downstream tasks like 3D reconstruciton. However, generative turntable video (GenTTV) technology is still in its infancy, and progress in general video synthesis has not directly translated into production-ready, object-centric rotational videos. Both commercial tools~\cite{runwaygen2,lumaai} and recent research on controllable video generation~\cite{Chen2023ControlAVideoCT,voleti2024sv3d, Gu2025DiffusionAS,zhang2023ControlNet} still face challenges in maintaining strict SKU identity, preserving metric scale, enforcing view-consistency across a 360-degree orbit, and handling difficult materials such as high-gloss or translucent surfaces.


The challenges for GenTTV are multifaceted, with the core issue being the extremely high demand for geometric and photometric consistency. Minor content drift or jitter that might be tolerable in narrative or stylized videos becomes a critical flaw in a turntable sequence, breaking the 3D illusion and violating the strict visual stability required in product visualization scenarios. Furthermore, an ideal GenTTV must not only serve as a visual preview but also be reconstructable, meaning its consistency must be high enough to serve as a reliable data source for downstream 3D tasks like baking into seamless UV texture maps~\cite{huang2025mv,ravi2020pytorch3d} or training neural rendering models such as 3D Gaussian Splatting and Neural Radiance Field~\cite{kerbl20233d, mildenhall2021nerf}. This requires the generation process to maintain high fidelity to camera intrinsics and extrinsics, radiometry, and surface material properties~\cite{walter2007microfacet}, but existing general-purpose video generation models are not optimized for these strict 3D constraints. As a result, reconstructions from their outputs often exhibit severe artifacts, especially around fine, thin structures or complex materials such as fur, fabrics, and specular or transparent surfaces.


In this paper, we introduce TAPESTRY, a framework for generating high-fidelity turntable videos with strict geometric consistency. We adopt a strongly geometry-constrained paradigm: given a 3D mesh, we render precise geometric features such as normal and position maps into a control video and feed it as pixel-level guidance to a modern video diffusion model. This design reframes the generative task from generic video synthesis to performing precise “visual texturing” on a fixed geometric scaffold, which fundamentally stabilizes the structure and appearance across views. To achieve complete surface coverage, we further design a multi-stage generation pipeline with 3D-aware inpainting: by rotating the model and using already generated content as context in a second pass, the method fills regions that are self-occluded in the initial views. 
Furthermore, the strict multi-view consistency allows our generated TTVs to serve as robust training signal for 3D Gaussian Splatting (3DGS)~\cite{kerbl20233d}, enabling the creation of photo-realistic, interactive 3D web viewers directly from our video outputs.
Our method not only makes the generated turntable video a valuable, directly viewable digital asset, but also provides a robust, high-fidelity data source for subsequent generation of neural representations~\cite{mildenhall2020nerf,  muller2022instant, kerbl20233d, wang2021neus} or seamless UV texture maps~\cite{laine2020nvdiffrast}, ultimately yielding a complete, high-quality 3D asset from an initially untextured model. Demonstrating its accessibility, the entire framework is trained on a single DGX\cite{nvidia_dgx_spark_2025} Spark supercomputer, showcasing its feasibility within a low-budget setting.


In summary, our main contributions are:
\begin{itemize}
    \item We introduce TAPESTRY, a novel framework for generating turntable videos with strong geometric consistency by conditioning video diffusion models on explicit 3D mesh geometry.
    \item We propose a novel progressive pipeline with context-aware inpainting that synthesizes complete and seamless textures from our generated turntable videos. 
    \item  We demonstrate the efficiency and versatility of our method: trainable on a single DGX Spark, TAPESTRY produces turntable videos with sufficient consistency to directly drive 3DGS reconstruction, enabling high-quality interactive 3D visualization.
\end{itemize}

%% file: sec/2_related_work.tex
\section{Related Work}
\label{2_related}

\begin{figure}[t]
\begin{center}
\includegraphics[width=\linewidth]{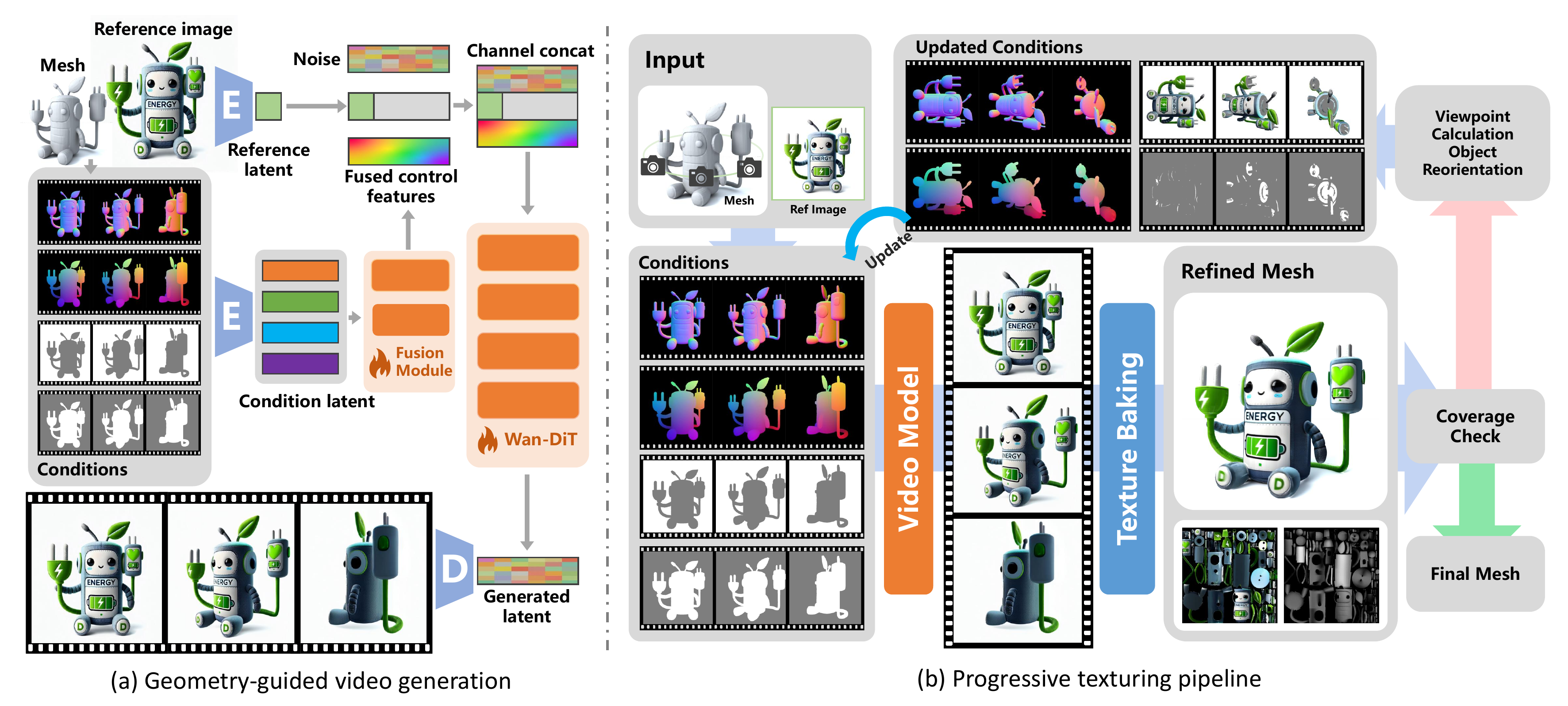}
\end{center}
\caption{An overview of the TAPESTRY architecture. \textbf{(a)} Geometry-guided video generation. Our method generates a 3D consistent Turntable Video by injecting multi-modal geometric conditions and reference context into a DiT-based video diffusion model. \textbf{(b)} Our progressive texturing pipeline. We iteratively generate TTVs from new, optimized viewpoints and fuse their projections via Texture Baking. Each pass is conditioned on previously generated textures to ensure global consistency, continuing until full coverage is achieved.}
\label{Overview}
\end{figure}

\textbf{Turntable Video Productions.}
Turntable videos (TTVs) are an industry standard for object visualization. Traditionally, their production relies on either labor-intensive physical photography or meticulous digital rendering in CG software~\cite{blender,maya}. The latter, though a digital process, still requires significant artistic effort in scene setup, material authoring~\cite{pharr2023physically}, and multi-point lighting to achieve realism, accompanied by lengthy rendering times. The high cost of both methods has spurred the exploration of generative approaches.

\textbf{Multi-view and Video Generation.}
Generative approaches to this challenge have largely followed two main paths. Before the maturation of video models, a popular approach was to synthesize a static set of multi-view images to represent a 3D object. Many Methods have demonstrated the ability to generate plausible, varied views from text~\cite{Liu2023SyncDreamerGM,huang2024dreamcontrol} or a single image~\cite{wu2023hyperdreamer,Liu2023Zero1to3ZO,liu2023one,shi2023mvdream} using powerful 2D diffusion priors. The primary limitation of this approach is consistency. As each view is generated with a degree of independence, ensuring photometric and geometric coherence is notoriously difficult.

To overcome the consistency limitations of static images, leveraging the temporal coherence of video models is a natural evolution. The advent of large-scale video diffusion models ~\cite{NEURIPS2022_39235c56, Blattmann2023StableVD, Chen2023VideoCrafter1OD,yang2024cogvideox, videoworldsimulators2024, kong2024hunyuanvideo, wan2025} has opened new avenues for dynamic content creation. Inspired by ControlNet~\cite{zhang2023ControlNet,chong2024T2IAdapter, Chen2023ControlAVideoCT, chen2024pixartdeltafastcontrollableimage} for images, numerous controllable video synthesis methods now allow conditioning on explicit signals like depth maps~\cite{Chen2023ControlAVideoCT,Alhaija2025CosmosTransfer1CW, Jiang2025VACEAV}, canny edges~\cite{Alhaija2025CosmosTransfer1CW, Jiang2025VACEAV}, 3D positions~\cite{Zhang2024WorldconsistentVD}, tracking video~\cite{Gu2025DiffusionAS}. Although these methods provide enhanced structural control, they often only achieve visual plausibility and fail to meet the strict, long-range, and pixel-perfect consistency required for a full 360-degree TTV. TAPESTRY builds upon this foundation and specializes it for this challenging long-range task.

\textbf{Texturing from Generative Views.}
Beyond providing fast, high-fidelity previews of objects, a primary application of generative views lies in downstream texturing and 3D reconstruction tasks. Early methods employed score distillation sampling (SDS) to optimize implicit representations such as NeRF~\cite{mildenhall2020nerf,poole2022dreamfusion, lin2023magic3d}, NeuS~\cite{wang2021neus,chen2023fantasia3d, wang2023prolificdreamer}, and their variants. However, suffering from computational inefficiency and perceptual biases inherent in image models, these approaches often exhibit the multi-head problem and slow convergence~\cite{wang2023prolificdreamer}. 
Subsequent works discovered the capability of pre-trained image models to generate multi-view consistent images, leading to approaches that either reconstruct implicit representations from multi-view images~\cite{liu2023one2345, Liu2023Zero1to3ZO, shi2023mvdream} or directly map them to mesh UV coordinates like SyncMVD~\cite{Liu2023TextGuidedTB}, MV-Adapter~\cite{huang2025mv}, and HY3D-2.0~\cite{Zhao2025Hunyuan3D2S}. However, these methods suffer from difficult-to-resolve artifacts in self-occluded regions—seams and color inconsistencies—due to limited viewpoints and inherent model constraints.

In contrast, a dense set of target views (TTV) provides continuous surface coverage, fundamentally mitigating the ambiguities and information loss inherent in sparse-view approaches. The high-fidelity, geometrically-consistent TTVs generated by TAPESTRY provide an ideal input for downstream tasks—whether for traditional texture back-projection, benefiting from reduced seams and artifacts, or for training neural implicit representations such as 3D Gaussian Splatting~\cite{kerbl20233d, muller2022instant}, where input density and consistency directly translate to higher fidelity and more refined final assets.

%% file: sec/3_method.tex
\section{Method}
\label{sec_method}
Our approach addresses the challenge of automatically generating a high-fidelity appearance for an untextured 3D mesh. We reframe this task as a geometry-conditioned video generation problem, proposing to synthesize a 360-degree turntable video (TTV) of the object as a high-quality intermediate representation. The central technical challenge, therefore, is to generate this TTV with strict 3D consistency, overcoming the content drift and structural distortion common in existing video models. Our core idea is to use explicit geometric features to strictly constrain a powerful video diffusion model for this purpose.


To detail the effectiveness of our method, this section is divided into two main parts. First, in Sec. ~\ref{sec_method_video}, we will provide an in-depth introduction to our core method for Geometry Guided 3D Consistent Video Generation, including the construction of its multi-modal conditions and its latent space injection mechanism. Subsequently, using the generated TTV as the core intermediate representation, we introduce a progressive texturing pipeline to create a complete asset Sec.~\ref{sec_method_texture}, which incorporates a novel 3D-Aware Inpainting process to address self-occlusion issues. 


\subsection{Geometry-guided Video Generation} 
\label{sec_method_video}
\textbf{Multi-view Geometric Feature Condition.}
The core of our method is a novel multi-modal geometric conditioning framework designed for a video diffusion model. By introducing a set of meticulously designed conditions containing strong geometric priors into a pre-trained video model and performing full-finetuning, this framework achieves a tight coupling between the generated video content and the 3D structure. Our approach divides this process into two main stages: (1) multi-modal condition preparation, and (2) geometry-guided latent space injection. The overall architecture is illustrated in Fig.~\ref{Overview}.

\textbf{Multi-modal Condition Preparation.}
To provide the model with comprehensive and unambiguous guidance for generating a high-quality, 3D-consistent TTV, we prepare both geometric conditions for pixel-level structural constraints and a reference condition for high-level content and style guidance.
For the geometric conditions, we first define a smooth virtual camera trajectory orbiting the given normalized 3D mesh. In our experiments, we primarily use a standardized circular orbit path, where the position $\mathcal{O}(t)$ at time $t$ is defined as:
\begin{equation} 
\label{eq:orbit_path}
\mathcal{O}(t) = \left( r \cos\left(\frac{2\pi t}{T}\right), r \sin\left(\frac{2\pi t}{T}\right), z \right),
\end{equation}
where $r$ is the orbit radius, $z$ is the camera height, and $T$ is the total number of frames. At each position, the camera is oriented to look at the origin. After uniform sampling $T$ camera poses, we render a series of strictly aligned geometric feature videos. The two most important are the Normal Video, which encodes surface orientation, and the Position Map Video, which provides world-coordinate information. The Normal Video provides fine-grained, local surface details, crucial for high-fidelity texture synthesis, while the Position Map Video offers a global, absolute spatial reference, which is vital for preventing long-range drift and maintaining structural integrity across the entire 360-degree rotation.

\textbf{Geometry-Guided Latent Space Injection.}
Regarding the mechanism for efficient fusion and synergistic injection of multi-modal conditions in the latent space, we first use the frozen encoder of a pre-trained lightweight video VAE~\cite{BoerBohan2025TAEHV} to map all prepared video-form conditions such as the normal video, position map video, a known video and an inpainting mask—individually into the latent space. Subsequently, we concatenate these geometric latents from different sources along the channel dimension and feed them into a small geometry fusion module of our design. This network, composed of several 2D convolutional layers and residual blocks, efficiently fuses the concatenated high-dimensional geometric features into a unified and information-dense geometric condition latent $\mathbf{c_\text{geo}}$. This approach is considered to be both lightweight and efficient while retaining maximum information, which will be validated in our experiments.

This geometric condition latent is then injected into the denoising process of the Diffusion Transformer (DiT)~\cite{Peebles2022ScalableDM} backbone. Specifically, the input to the DiT is augmented by concatenating the standard noisy latent with our geometric condition latent along the channel dimension. Additionally, a reference frame latent is also concatenated. This is generated by encoding the single initial frame and padding it across the frame dimension to match the video's length. This channel-wise concatenation forces the model to adhere to the geometric structure at a pixel level.
Meanwhile, for high-level semantic control, the input text prompt and the initial frame are converted into embeddings using pre-trained text and vision encoders (e.g., umT5~\cite{Chung2023UniMaxFA} and CLIP~\cite{Radford2021LearningTV}). These embeddings are stitching along the sequence to get context embedding and fed into the cross-attention layers of the DiT blocks as context. This conditioning mechanism allows our model to generate a highly consistent TTV that strictly follows the underlying 3D geometry while simultaneously adhering to the desired text description and reference style.

\begin{figure}[t]
\begin{center}
    \includegraphics[width=\linewidth]{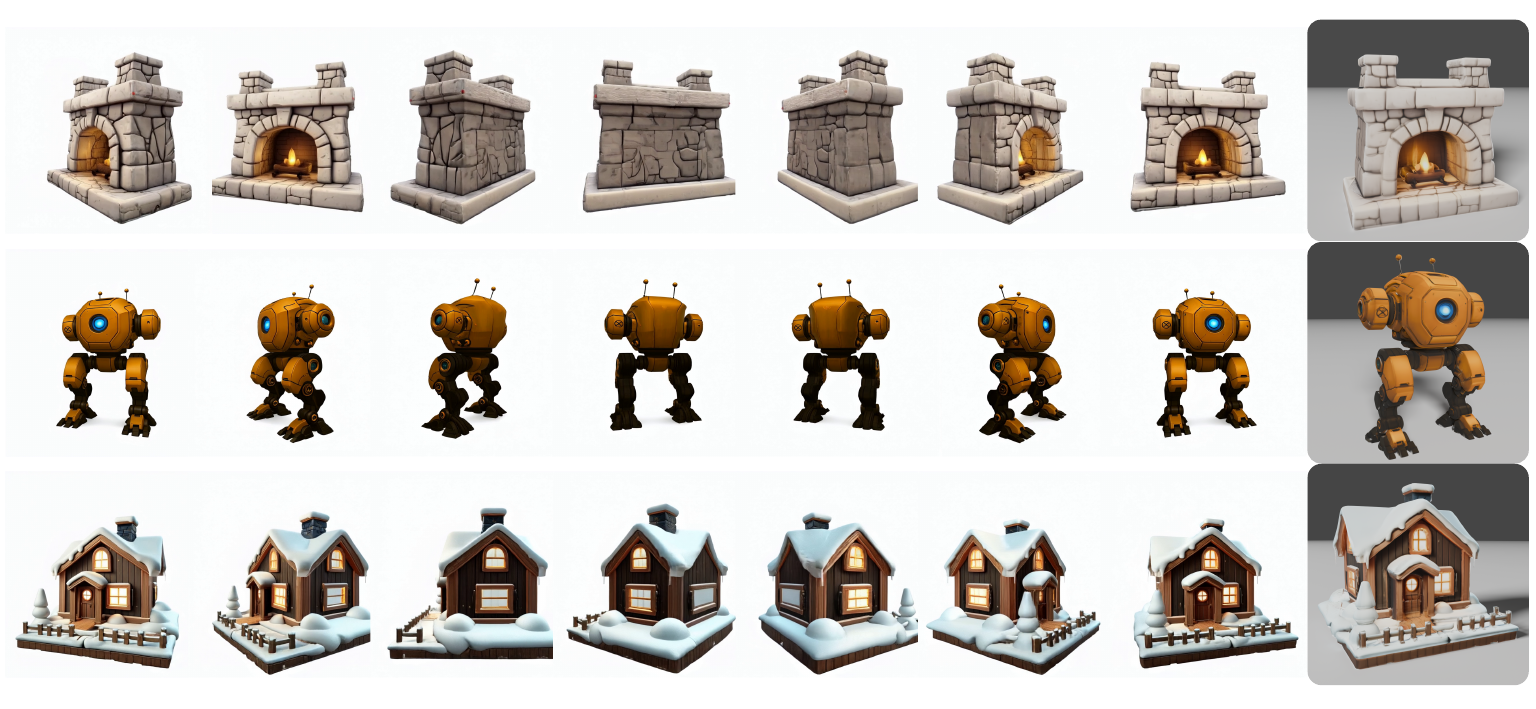}
\end{center}
\caption{Qualitative results of TAPESTRY. Our method generates a consistent Turntable Video, from which a complete and high-fidelity textured asset is produced.}
\label{fig:gallery}
\end{figure}

\subsection{High-Fidelity Texturing from Video}
\label{sec_method_texture}
Leveraging the ability to generate 3D-consistent video sequences, we apply our method to a highly challenging downstream task: automatically generating high-quality texture maps for arbitrary 3D models. As texture mapping places extremely stringent demands on 3D consistency, any minor inconsistency will starkly manifest on the final 3D asset in the form of seams, ghosting, or blurring.
However, directly projecting the video from a single generation pass back onto the model's surface often results in an incomplete texture map due to self-occlusion or the camera's failure to cover all regions. To address this problem, we design a progressive texturing pipeline that incorporates a novel 3D-Aware Inpainting mechanism. 

\textbf{Multi-view Texture Projection.}
We back-project pixel information from each frame onto the mesh's UV space using ray tracing to confirm mapping relationships and visibility. For weighted blending, we introduce an angle incentive term to prioritize surfaces perpendicular to the camera view: $\lambda_{\text{angle}}=\cos\theta^4$, and a depth gradient penalty term to mitigate edge misalignment: $\lambda_{\text{depth}}=\mathrm{clip}(\nabla^2 d, 0, 1)$. The similarity mask is defined as $S_i=\lambda_{\text{angle}}(1-\lambda_{\text{depth}})$. We then back-project the similarity mask and the original images according to the ray tracing results, yielding a partial texture map and a partial weight map. By performing a weighted summation of the back-projection results from all images, we obtain an initial texture map and an accumulated corresponding confidence map. 

\textbf{3D-Aware Inpainting and Progressive Refinement.}
To achieve complete surface coverage and handle complex self-occlusions, we design a progressive texturing pipeline built upon our 3D-Aware Inpainting mechanism.
To fill the holes in the partial texture map, we must generate views of the previously occluded regions. A naive solution would be to generate a second, independent TTV from a different angle. However, this approach would reintroduce the very problem we aim to solve: inconsistency. Since the two generation processes are separate, the model would likely produce conflicting appearances for overlapping regions, destroying the global 3D consistency of the asset.

To overcome this, our pipeline systematically refines the texture through iterative, context-aware generation passes. Each iteration begins by algorithmically determining an optimal base rotation for the object itself, chosen to maximize the visibility of untextured regions. Crucially, while the object is reoriented, the camera trajectory remains identical to the initial 360-degree orbit.
With the object in its new base rotation, the critical step of condition preparation follows. A comprehensive set of conditioning videos is rendered along the fixed camera path. This set includes not only the geometric information corresponding to the rotated object, but most importantly, a Partial Texture Video and an Inpainting Mask Video, which are also rendered from the rotated, partially-textured mesh. The Partial Texture Video provides strong contextual priors of existing content, while the Inpainting Mask explicitly directs the model to fill the missing regions. With these conditions, the model performs a context-aware completion, ensuring the newly generated content seamlessly matches the existing appearance. The resulting TTV is then projected back to update the master texture map. This process can be repeated with different object rotations until the surface is fully covered.

\textbf{Multi-Stage Texture Fusion.}
For complex objects requiring complete coverage, we define multiple camera trajectory sets and pre-compute their contributions on confidence maps to determine viewpoint order. We then perform iterative video generation and material fusion. At each stage, the Refined TTV is converted to partial texture and confidence maps via weighted back-projection. To obtain a seamless high-fidelity texture, we fuse the two partial sets through weighted summation, where weights derive from their respective confidence maps. This weight-based blending, combined with the inpaint module's consistency constraints, ensures that more reliable viewpoints contribute more significantly to the final result. The confidence maps are subsequently updated for use in the next Texture Refinement iteration.

%% file: sec/4_experiments.tex
\section{Experiments}
\label{sec_exp}
In this section, we systematically evaluate TAPESTRY through comprehensive experiments. We first showcase our generated 3D-consistent Turntable Videos (TTVs) and the final textured assets in Fig.~\ref{fig:gallery}, demonstrating our ability to achieve both 3D consistency and high-fidelity appearance. We then present detailed quantitative and qualitative comparisons for our two key outputs: the TTVs as an intermediate representation, and the final textured 3D assets. 

\subsection{Implementation Details}
Our implementation is built upon the VideoX-Fun\cite{videoxfun} framework, using the open-source Wan2.1-Fun-V1.1-Control model as the base for fine-tuning. We adopt LoRA\cite{Yu2023LORA} with rank set to 128 for parameter-efficient adaptation, where all linear layers within the DiT blocks and the proposed Geometry Control Mix module are optimized. We further apply the LoRA+\cite{hayou2024LOARPLUS} strategy, setting a base learning rate of $1\times10^{-5}$, and scaling the learning rates of the LoRA matrices B and our proposed Geometry Control Mix module by a factor of 4.
Training is conducted on a single DGX Spark~\cite{nvidia_dgx_spark_2025} supercomputer and takes approximately 10 days. Leveraging our lightweight implementation, generating a single 61-frame video takes only 30 seconds on the DGX system when using 4-step inference. Code and model weights will be released.

\textbf{Dataset.} 
For training, we construct a large-scale, diverse dataset of synthetic turntable videos. We start with approximately 30K high-quality textured 3D assets from various categories, collected from Objaverse~\cite{deitke2023objaverse}. During data preprocessing, each 3D model is normalized to fit within a unit sphere and centered at the origin. For each model, we render a 61-frame turntable video in the Blender Eevee engine, following the circular trajectory defined in Sec.~\ref{sec_method_video}. To enhance viewpoint diversity, the camera's radius r and height z are randomly sampled within a fixed range. We also compute an optimal field of view (FOV) based on the camera distance and the object's bounding box to maximize the proportion of informative pixels in the video, preventing the model from overfitting to specific camera parameters. To further improve generalization and avoid pose bias, we apply a random initial rotation to each object before rendering. To achieve realistic lighting, we perform data augmentation using 4 common environment maps, while also randomizing the light source intensity and angle within a certain range. This process ultimately yields a dataset of approximately 120K turntable videos.


\begin{figure}[htbp]
    \centering
    \includegraphics[width=\linewidth]{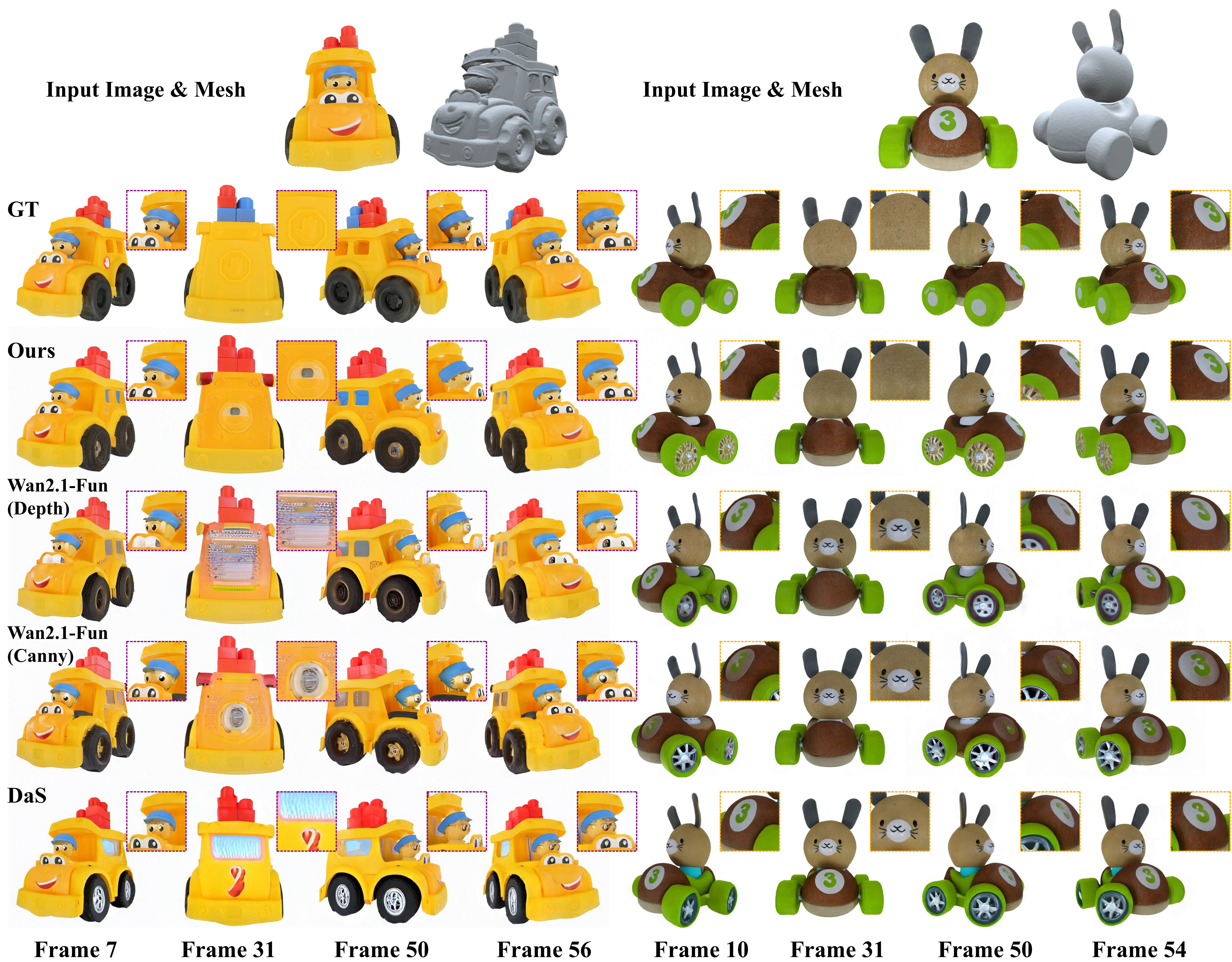}
    \caption{Qualitative comparison. As shown, the controllable video baselines suffer from significant appearance drift, the Janus problem. In contrast, our method maintains strong object identity and consistency throughout the rotation, effectively eliminating such artifacts.}
    \label{fig:comparison} 
    \includegraphics[width=\linewidth]{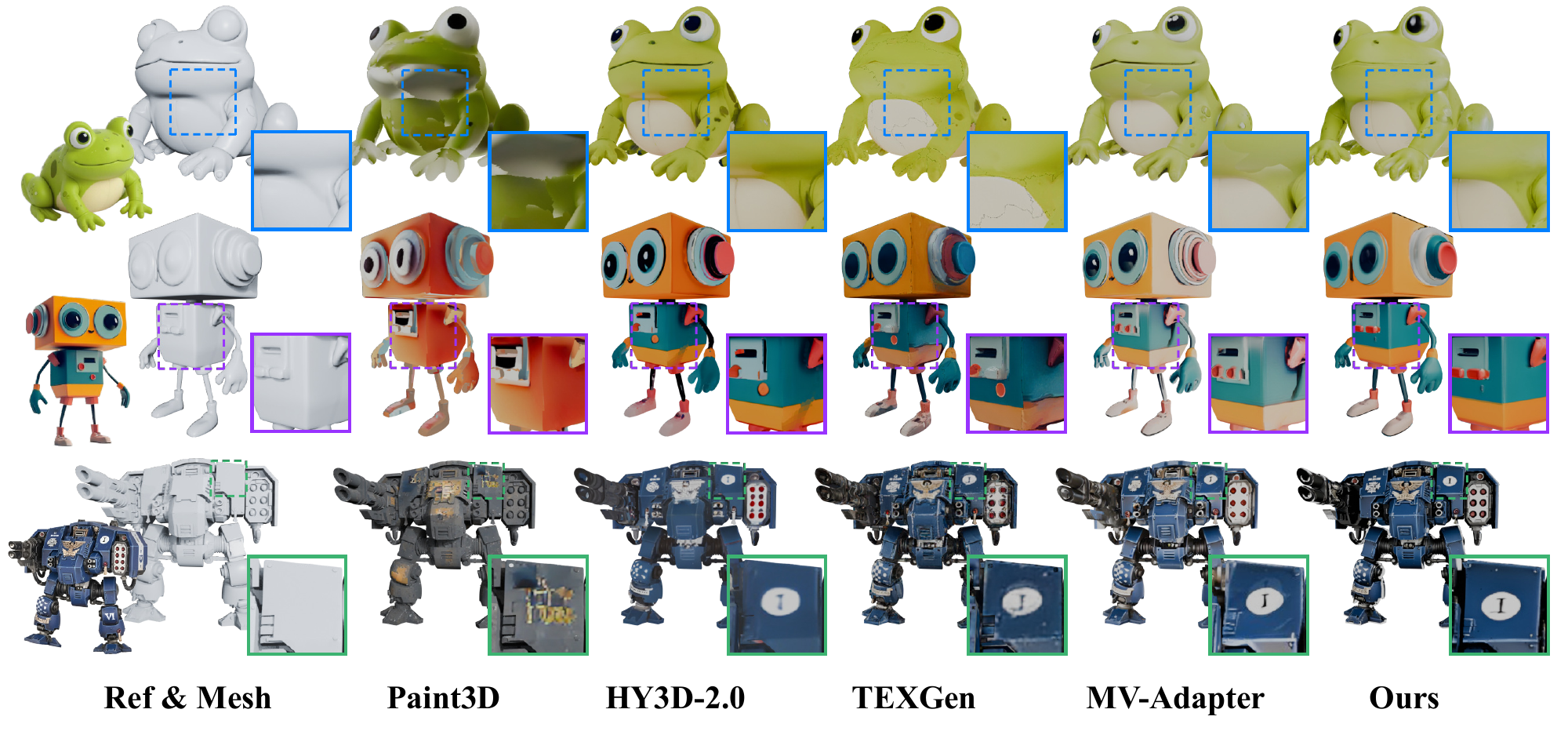}
    \caption{Qualitative comparison on textured mesh. The first two rows show inconsistency issues in self-occluded regions, where existing methods produce noticeable seams due to insufficient inter-view smoothness. The third row reveals consistency deficiencies, where weak multi-view constraints cause blurring (HY3D-2.0) and ghosting (MV-Adapter) in high-frequency textures.}
    \label{fig:texture_comparison} 
\end{figure}

\subsection{Comparisons}
\textbf{Consistent Video Generation.} 
We evaluate the quality and 3D consistency of the Turntable Videos generated by TAPESTRY on 200 meshes randomly sampled from the GSO benchmark~\cite{downs2022google} and the Objaverse~\cite{deitke2023objaverse} testset. While general controllable video models excel at plausible motion, they are not optimized for the long-range, rigid-body consistency demanded by a 360-degree TTV, often resulting in noticeable artifacts and temporal jitter. We compare TAPESTRY against two SOTA approaches: a representative controllable video generation framework\cite{videoxfun} conditioned on depth and Canny edges respectively, and Diffusion as Shader(DaS)~\cite{Gu2025DiffusionAS}.

Quantitatively, we use several standard metrics to evaluate the generated TTVs against the ground-truth videos. For these, we compute PSNR\cite{wang2004image}, SSIM\cite{hore2010image}, LPIPS\cite{zhang2018unreasonable}, and FVD\cite{unterthiner2018towards} between the generated videos and the ground-truth videos. As shown in Tab.~\ref{tab:video_eval}, our results are significantly better than all baselines, which strongly validates its superior ability in maintaining geometric consistency and video quality.

The qualitative results are shown in Fig.~\ref{fig:comparison}. The different conditioning methods of the controllable video framework and Diffusion as Shader exhibit clear inconsistencies. Most typically, they suffer from the classic multi-head/Janus problem, which our method does not have. Additionally, their surface details flicker and drift during rotation. In stark contrast, the videos generated by TAPESTRY maintain perfect structural stability and appearance consistency throughout the 360-degree rotation, with clear details and no distortion. 
Furthermore, our user study results confirm the superiority of our method. When asked to choose the best result, users preferred TAPESTRY overwhelmingly across three dimensions: alignment with initial reference (79.8\% preference), geometric consistency (80.9\% preference), and overall video quality (76.1\% preference). This clearly demonstrates the effectiveness of our geometry-conditioning paradigm.
\begin{table}[tb]
\centering
\caption{Quantitative comparison of generated Turntable Videos. }
\setlength{\tabcolsep}{6pt}
\setlength{\tabcolsep}{1.5mm}{
\begin{tabular}{@{}l|cccc}
\toprule
\multicolumn{1}{c|}{\textbf{}}      & \multicolumn{4}{c}{\textbf{Video Metrics}}                      \\
\multicolumn{1}{l|}{\textbf{Method}} & PSNR~(↑)  & SSIM~(↑)  & LPIPS~(↓) & FVD~(↓)  \\ \midrule
\ Wan2.1-Fun (Depth)                  & 23.48          & 0.904          & 0.084          & 277.9   \\
\ Wan2.1-Fun (Canny)                  & 24.32          & 0.916          & 0.074          & 235.0     \\
\ Diffusion as Shader                 & 23.21          & 0.906          & 0.090          & 295.7       \\
\ {\bf Ours}                       & {\bf 25.79} & \textbf{0.924} & \textbf{0.066} & \textbf{189.9}  \\ \bottomrule
\end{tabular}
}
\label{tab:video_eval}
\end{table}

\textbf{Texture Generation.}
We further evaluate the quality of the final texture maps generated by our progressive pipeline. The ultimate goal is to produce a complete, seamless, and high-fidelity UV map that is free from artifacts caused by viewpoint inconsistencies or self-occlusion. We compare our final textured assets against several state-of-the-art image-to-texture methods, including Paint3D~\cite{zeng2024paint3d}, Hunyuan3D-2.0 (HY3D-2.0)~\cite{Zhao2025Hunyuan3D2S}, TEXGen~\cite{brown2022texgen}, and MV-Adapter~\cite{huang2025mv}.



Following~\cite{zeng2024paint3d,huang2025mv,Zhao2025Hunyuan3D2S}, we quantitatively evaluate the FID, KID($\times 10^4$) and CLIP score on multi-view renderings of the generated textures in the constructed testset. As reported in Tab.~\ref{tab:tex_eval}, TAPESTRY achieves a substantially better FID, KID($\times 10^4$) and CLIP score. We also report the average runtimes in Tab.~\ref{tab:tex_eval}. While slower than baseline texture generation methods, the cost is justified by our dual outputs: TAPESTRY delivers both superior quality textures and geometry-aligned TTVs. Since the TTV serves as a valuable, standalone appearance representation, this comprehensive generation renders the runtime an acceptable trade-off.

Figure~\ref{fig:texture_comparison} presents the qualitative comparison on several challenging models. Paint3D fails on complex UV layouts (third row). TEXGen shows unacceptable seams with complex UVs (first row inset). HY3D-2.0 produces inconsistent multi-view images, causing color discontinuities at self-occluded regions and unclear textures. While MV-Adapter improves consistency, it still suffers from color discrepancies in self-occluded areas. Both HY3D-2.0 and MV-Adapter exhibit ghosting or blurring (third row), showing that orthogonal six-view projections cannot maintain pixel-level consistency in complex regions. In contrast, our method generates consistent, seamless, and sharp results across all challenging areas, demonstrating the advantages of TTV.

\begin{table}[t]
\centering
\caption{Quantitative comparison of generated textures. All metrics are evaluated on multi-view renderings.}
\setlength{\tabcolsep}{6pt}
\setlength{\tabcolsep}{1.5mm}{
\begin{tabular}{@{}l|cccc}
\toprule
\multicolumn{1}{c|}{\textbf{}}      & \multicolumn{4}{c}{\textbf{Texture Metrics}}                      \\
\multicolumn{1}{l|}{\textbf{Method}} & FID~(↓)  & KID~(↓)   & CLIP-S~(↑) & Time~(↓) \\ \midrule
\ Paint3D                  & 43.47          & 36.63          & 0.8880           & 96s \\
\ HY3D-2.0                  & 31.45          & 22.52          & 0.9165            & 33s  \\
\ TEXGen                 & 39.11          & 35.09          & 0.8906              & \textbf{15}s \\
\ MV-Adapter                 & 30.68          & 25.73          & 0.9291            & 32s  \\
\ \textbf{Ours}                       & \textbf{26.90} & \textbf{16.21} & \textbf{0.9488} & 87s \\ \bottomrule
\end{tabular}
}
\label{tab:tex_eval}
\end{table}


To validate the key components of the TAPESTRY framework, we conduct a series of ablation studies, focusing on two critical aspects: (1) the contribution of each modality within our geometric conditions, and (2) the effectiveness of our progressive texturing pipeline.

\textbf{Effect of Geometric Conditions.} 
Our core hypothesis is that both local and global geometric priors are necessary for generating high-quality, consistent TTVs. To test this, we train three model variants:  (a) w/o Normal, which removes the local surface detail cue, (b) w/o Position, which removes the global spatial reference and (c) Ours,using both Normal and Position map videos; As shown in Tab.~\ref{tab:ablation}, our full model achieves the best performance across all metrics. The qualitative results also reveal the distinct roles of each condition: the model using only position maps, while maintaining good global structural stability, produces mediocre surface details and lighting effects. Conversely, the model using only normal maps, while capable of generating rich details, occasionally exhibits slight structural drift when faced with complex geometries. Only the combination of both can simultaneously guarantee global geometric consistency and high-quality local details, thereby generating the most realistic videos.

\begin{figure}[t]
\begin{center}
    \includegraphics[width=\linewidth]{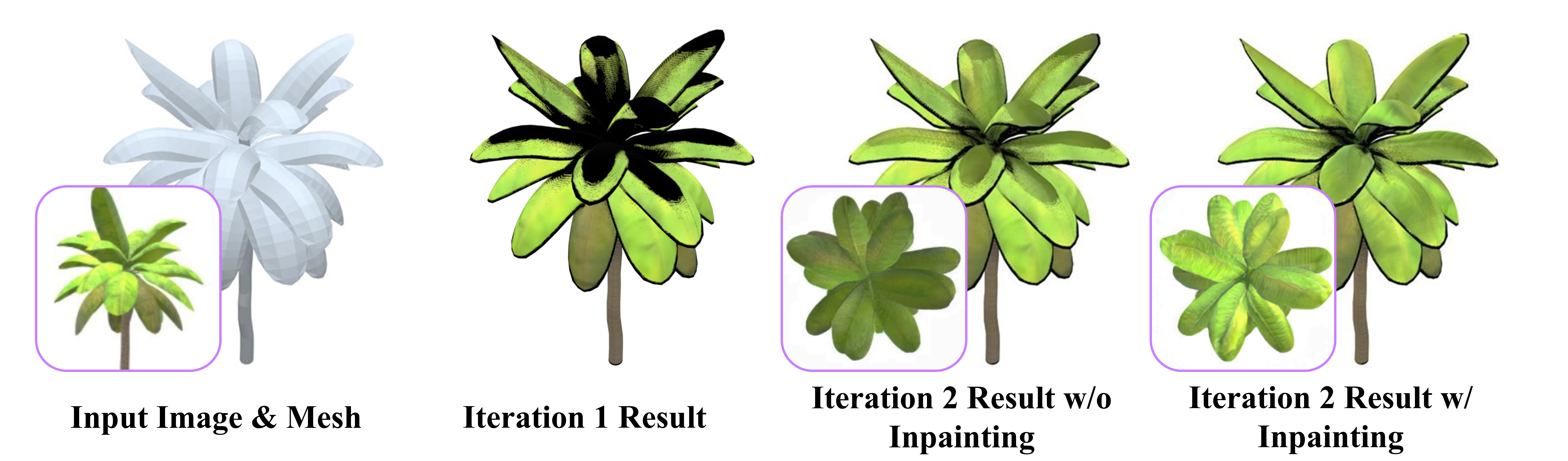}
\end{center}
\caption{The initial Iter.1 result has an incomplete texture due to self-occlusion. The w/o Inpainting baseline that generates a second TTV independently suffers from visible seams and color shifts.In contrast, our full pipeline with context-aware inpainting produces a seamless and globally consistent result. }
\label{fig:ablation}
\end{figure}

\textbf{Effectiveness of Progressive Texturing.} To demonstrate the superiority of our context-aware inpainting mechanism, we compare it against a "Naive Fusion" baseline. This baseline generates two independent TTVs from different base rotations and then simply blends the resulting partial textures. As shown in Fig.~\ref{fig:ablation}, the "Naive Fusion" method, due to its lack of context, produces visible seams and abrupt color shifts at the junctions of the independently generated content, as each generation pass creates a conflicting "imagination" of the appearance. In stark contrast, our TAPESTRY pipeline, by explicitly conditioning each subsequent pass on previously generated textures, ensures all new content is a consistent extension of the existing appearance, resulting in a seamless and globally coherent final texture map.

\begin{table}[t]
\centering
\caption{Quantitative ablation results on the Geometry Control Fusion module}
\setlength{\tabcolsep}{6pt}
\setlength{\tabcolsep}{1.5mm}{
\begin{tabular}{@{}l|cccc@{}}
\toprule
\textbf{Method} & PSNR~(↑)  & SSIM~(↑)  & LPIPS~(↓) & FVD~(↓) \\ \midrule
w.o normal      & 25.15                 & 0.918                 & 0.073                  & 203.8                \\
w.o position    &   25.66               & 0.921                 & 0.069                  & 191.1                \\
\textbf{Ours}      & \textbf{25.79} & \textbf{0.924} & \textbf{0.066} & \textbf{189.9}                \\ \bottomrule
\end{tabular}
}
\label{tab:ablation}
\end{table}

%% file: sec/5_discussion.tex
\section{Discussion}

\begin{figure}[t]
\begin{center}
    \includegraphics[width=0.85\linewidth]{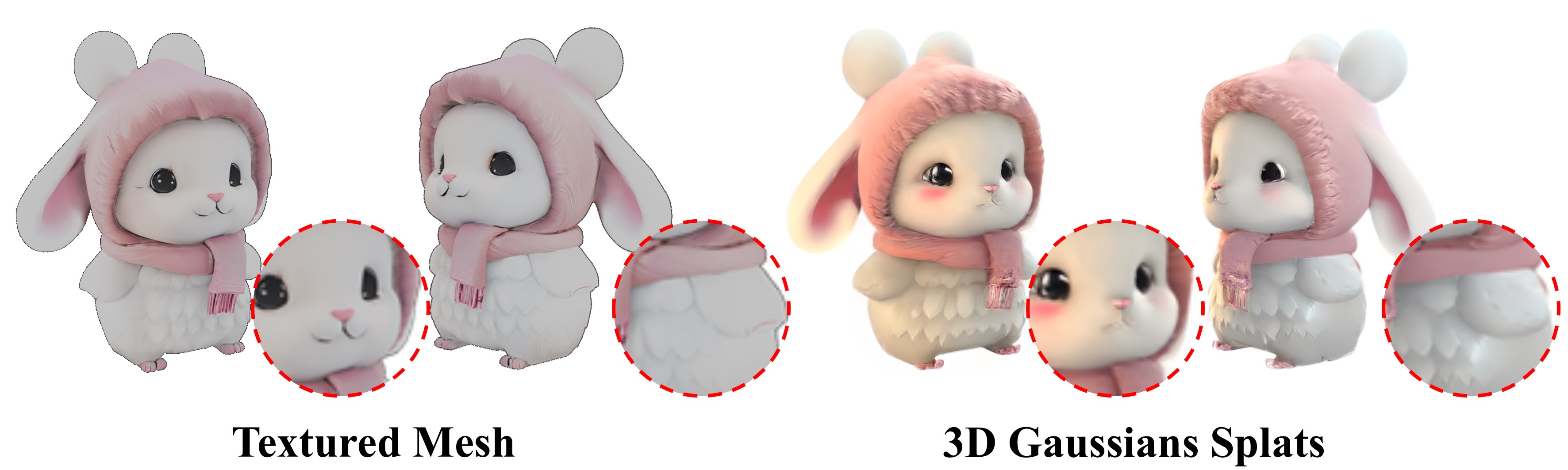}
\end{center}
\caption{Our consistent TTVs can supervise both traditional texturing and neural representations like 3DGS, with the latter better capturing volumetric details like fur.}
\label{fig:gaussians_splatting}
\end{figure}
\label{sec_discussion}

\textbf{Turntable video on Neural Assets.}
Beyond texturing meshes, our consistent TTVs serve as a robust training signal for neural assets like 3D Gaussian Splatting. As shown in Fig.~\ref{fig:gaussians_splatting}, this yields a high-fidelity representation that can capture complex volumetric details, such as fur, more realistically than a standard UV texture map. This demonstrates the versatility of our generated TTVs. Additional results and an interactive web demo are provided in the supplementary video.

\textbf{Conclusion.} 
We introduced TAPESTRY, a framework that generates high-fidelity 3D appearances by first synthesizing a geometrically-consistent Turntable Video. By injecting strong geometric priors into a video model, we ensure the TTV is structurally stable and free from drift. This high-quality TTV then serves as a robust intermediate for our progressive, context-aware inpainting pipeline, which produces complete and seamless UV textures. Experiments show our method significantly outperforms state-of-the-art approaches in generating consistent and detailed 3D assets, highlighting its potential for automated content creation.
Furthermore, we demonstrate the practical value of TAPESTRY by training it on a single DGX Spark and leveraging its high-consistency outputs to directly drive 3DGS for interactive web visualization.

\textbf{Limitation.} 
Despite the encouraging results, our method still has several limitations.
First, the performance of TAPESTRY is highly dependent on the quality of the input 3D mesh. A clean mesh with a well-defined topology is a prerequisite for achieving high-fidelity results. 
While our method maintains global consistency even on suboptimal inputs, the final output quality is inevitably bounded by the input geometry.
Second, the generated appearance contains lighting from the video model, rather than from a controllable source like an environment map, which hinders re-lighting.